\documentclass[letterpaper, 10 pt, conference]{ieeeconf}  % Comment this line out if you need a4paper [letterpaper]

\IEEEoverridecommandlockouts                              % This command is only needed if 
\usepackage{xcolor}
\definecolor{gt}{RGB}{128,128,128}      % gray for Ground Truth
\definecolor{llama}{RGB}{220,220,220}   % gray for VideoLLaMA / Qwen
\definecolor{llava}{RGB}{200,220,255}   % light blue for LLaVA
\definecolor{ours}{RGB}{255,220,200}    % light orange for DriveSafe
\definecolor{good}{RGB}{0,150,70}       % soft green text
\definecolor{bad}{RGB}{200,50,50}       % soft red text

\usepackage{graphicx}
\usepackage{subcaption}
\usepackage[font=small,labelfont=bf]{caption}
\usepackage{mwe}
\usepackage{amsmath}
\usepackage{tablefootnote}
\usepackage{url}
\usepackage{listings}
\usepackage{float}
\usepackage{multirow}
\usepackage{hhline} % For double lines
\usepackage{tablefootnote}
\usepackage{dingbat}
 \usepackage[table]{xcolor}
\usepackage[dvipsnames]{xcolor}
\usepackage{tikz}
\usepackage{bbding}
\usepackage{color}
 \usepackage[table]{xcolor}
\usepackage{cuted}
\usepackage{capt-of}
\usepackage{makecell}
\usepackage{algorithm}
\usepackage{algorithmic}
\usepackage{graphicx}
\usepackage{subcaption} % in preamble

\definecolor{cvprblue}{rgb}{0.21,0.49,0.74}
\usepackage{arydshln}
\usepackage[pagebackref,breaklinks,colorlinks,citecolor=cvprblue]
{hyperref}
\usepackage{array} % For customizi
\usepackage[most]{tcolorbox}
\usepackage[T1]{fontenc}
\usepackage{booktabs} % For professional looking tables
\usepackage[table]{xcolor} % For row colors
\usepackage{booktabs}
\usepackage{pifont}
\usepackage{graphicx}
\usepackage{mdframed} % in preamble
\usepackage{amsmath}
\usepackage{float}
\usepackage[capitalize]{cleveref}
\usepackage{amssymb}
\usepackage{adjustbox}

\newcommand{\cmark}{\ding{51}}%
\newcommand{\xmark}{\ding{55}}%

\definecolor{gt}{RGB}{235, 255, 235}      % gray for Ground Truth
\definecolor{llama}{RGB}{220,220,220}   % gray for Qwen
\definecolor{llava}{RGB}{200,220,255}   % light blue for LLaVA
\definecolor{ours}{RGB}{255,220,200}    % light orange for DriveSafe

\definecolor{good}{RGB}{0,150,70}       % soft green text
\definecolor{bad}{RGB}{200,50,50}       % soft red text

\newcommand{\resultbox}[2]{%
  \colorbox{#1}{\begin{minipage}[t][2.5cm]{\linewidth}
  \small #2
  \end{minipage}}%
}

\title{\LARGE \bf
DriveSafe: A Framework for Risk Detection and Safety \\Suggestions in Driving Scenarios
}

\author{Sainithin Artham, Shankar Gangisetty, Avijit Dasgupta, C. V. Jawahar% <-this % stops a space
\thanks{IIIT-Hyderabad, India
{\tt\small sainithin.artham@gmail.com, 
\{shankar.gangisetty@ihub-data., avijit.dasgupta@research., jawahar@\} iiit.ac.in}} %
}

\begin{document}
\maketitle

\begin{abstract}
Comprehensive situational awareness is essential for autonomous vehicles operating in safety-critical environments, as it enables the identification and mitigation of potential risks. Although recent Multimodal Large Language Models (MLLMs) have shown promise on general vision–language tasks, our findings indicate that zero-shot MLLMs still underperform compared to domain-specific methods in fine-grained, spatially grounded risk assessment. To address this gap, we propose DriveSafe, a framework for risk-aware scene understanding that leverages structured natural language descriptions. Specifically, our method first generates spatially grounded captions enriched with multimodal context—including motion, spatial, and depth cues. These captions are then used for downstream risk assessment, explicitly identifying hazardous objects, their locations, and the unsafe behaviors they imply, followed by actionable safety suggestions. To further improve performance, we employ caption–risk pairings to fine-tune a lightweight adapter module, efficiently injecting domain-specific knowledge into the base LLM. By conditioning risk assessment on explicit language-based scene representations, DriveSafe achieves significant gains over both zero-shot MLLMs and prior domain-specific baselines. Exhaustive experiments on the DRAMA benchmark demonstrate state-of-the-art performance, while ablation studies validate the effectiveness of our key design choices. Project page: \url{https://cvit.iiit.ac.in/research/projects/cvit-projects/drivesafe}.
\end{abstract}

\section{Introduction}
\label{Sec_Introduction}
% \textcolor{blue}{
% }

Risk assessment and safety prediction are central to many safety-critical domains such as aviation~\cite{Sikora2015Risk}, healthcare~\cite{Voskanyan2021Risk}, and robotics~\cite{8844289}, where anticipating hazards is essential for preventing catastrophic failures. Road transportation presents a similarly critical challenge: traffic accidents remain one of the leading causes of mortality worldwide, with an estimated $1.2$–$1.35$ million deaths annually and tens of millions of serious injuries~\cite{WHO2024}. In the United States alone, motor vehicle crashes claimed around $43,000$ lives in $2023$, corresponding to a fatality rate of about $12.2$ deaths per $100,000$ people~\cite{IIHS2023}. These sobering statistics highlight the urgent need for rigorous risk assessment and proactive safety interventions across all forms of driving. With the growing adoption of autonomous vehicles, this demand becomes even more pressing, as self-driving systems must not only perceive and plan effectively but also reason under uncertainty, anticipate hazards, and provide reliable safety suggestions in complex, real-world environments. 
%Central to this challenge is the ability to identify and reason about critical traffic agents, which forms the basis of risk-sensitive perception and prediction.

Existing works have approached this challenge from multiple perspectives: attention-based forecasting models such as RAIN~\cite{li2021rain} emphasize risk-aware trajectory prediction by highlighting salient agents, reinforcement learning frameworks with latent state inference~\cite{ma2021reinforcementlearningautonomousdriving} aim to capture hidden risk cues for decision-making, and interaction graph models~\cite{zhang2020interaction} are designed to represent spatio-temporal dependencies among on-road agents. While these approaches advance risk prediction, they do not address the problem of generating \textit{actionable safety suggestions}, which are crucial for practical deployment.

\begin{figure}[t]
    \centering
    \includegraphics[width=1\linewidth]{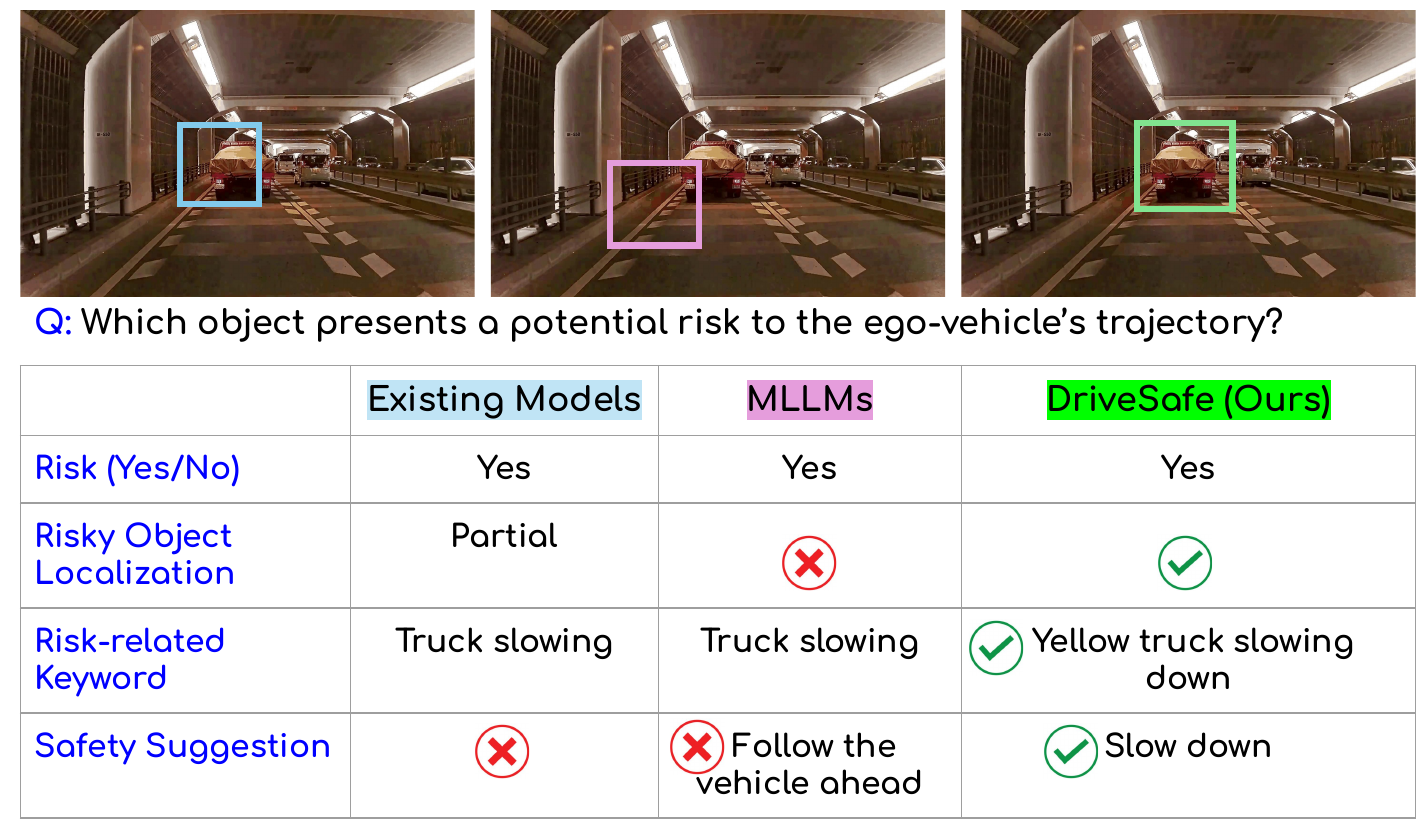}
    \caption{Previous works in driving scenarios~\cite{malla2023drama, bolya2022token, zhou2024hintspromptenhancingvisual} primarily address risk perception but fall short of offering actionable safety guidance. Similarly, general-purpose MLLMs~\cite{bai2025qwen25vltechnicalreport, li2024llavanextinterleavetacklingmultiimagevideo, zhang2025videollama3frontiermultimodal} are still unreliable in this regard. In contrast, our approach, \textit{DriveSafe}, integrates risk assessment with clear, human-understandable safety suggestions.}
    \label{fig:teaser}
\vspace{-.2cm}
\end{figure}

The curation of large-scale datasets such as DRAMA~\cite{malla2023drama} and Rank2Tell~\cite{sachdeva2023rank2tellmultimodaldrivingdataset} has further accelerated research in this direction. While a few works~\cite{zhou2024hintspromptenhancingvisual, ma2024videotokensparsificationefficient} have utilized these datasets, their primary focus has not been on advancing risk perception or prediction. As no existing dataset explicitly addresses safety suggestions, we extend the DRAMA~\cite{malla2023drama} dataset by explicitly associating critical objects with risk-related behavioral keywords (see Section~\ref{data-capture} for details).
%However, these methods focus primarily on risk assessment in isolation without leveraging the safety suggestion dimension. Moreover, the deployment of MLLMs in autonomous driving continues to face challenges such as hallucinated outputs and weak temporal alignment between video and language~\cite{Zhou2023Vision}, as we have demonstrated in this work.

On the other hand, MLLMs have recently shown strong performance in general video understanding tasks such as visual question answering~\cite{parikh2025roadsocialdiversevideoqadataset} and image captioning~\cite{liu2024improved}. General-purpose models like Qwen2.5-VL~\cite{bai2025qwen25vltechnicalreport}, Video-LLaMA3~\cite{zhang2025videollama3frontiermultimodal}, and LLaVA-Next~\cite{li2024llavanextinterleavetacklingmultiimagevideo} demonstrate impressive capabilities in generating natural language descriptions of visual scenes. However, these captions are typically generic and descriptive in nature, focusing on objects, activities, and context, rather than identifying potential hazards or issuing safety-related guidance. As a result, current MLLMs fall short in producing risk-aware or safety-critical captions that can inform proactive interventions in driving scenarios. Bridging this gap requires models that explicitly align visual understanding with safety reasoning, moving beyond surface-level descriptions toward actionable safety suggestions.

% %To address these gaps, 
% we first generate scene-level captions from driving videos using multimodal contextual cues such as optical flow, road and lane segmentation, and depth information. These captions are then provided to an LLM, which is prompted to produce risk assessment outputs. In addition to risk identification, we emphasize generating actionable safety suggestions that are directly grounded in the detected hazards\textit{ (ex: Slow down for Pedestrian nearby)}.
In this work, we take an alternative approach to risk assessment by avoiding direct fine-tuning of a MLLM. We argue that generic vision encoders as in MLLMs, while effective for broad video understanding tasks, are insufficient for safety-critical driving scenarios where subtle motion cues, spatial context, and depth perception play a decisive role. To address this, our method, \textit{DriveSafe}, leverages multimodal contextual signals—including optical flow, road and lane segmentation, and depth maps—together with video descriptions to generate driving-specific scene captions. These captions are then provided to a large language model (LLM), which is prompted to produce risk assessment outputs. Crucially, unlike prior approaches that focus solely on risk detection, DriveSafe explicitly generates actionable safety suggestions that are grounded in the detected hazards (e.g., “\textit{Slow down}”) (refer Fig.~\ref{fig:teaser}).

In summary, our key contributions are as follows:
\begin{enumerate}
    \item We propose \textit{DriveSafe}, a novel pipeline for driving risk assessment that avoids direct fine-tuning of MLLMs. Instead, it integrates multimodal contextual cues (optical flow, lane/road segmentation, depth) with video descriptions to produce driving-specific captions, which are then processed by an LLM for risk prediction and safety suggestion.
    
    \item We extend the DRAMA dataset by associating risky behaviors with critical objects, introducing safety-suggestion annotations. To ensure scalability and reproducibility, we annotate the training set automatically with an LLM while reserving manual annotations for the test set.
    
    \item Extensive experiments demonstrate that DriveSafe outperforms prior baselines in both risk assessment and safety suggestion quality, while maintaining strong generalization across diverse driving scenarios.
\end{enumerate}

\section{Related Works}
\label{realted-works}

\subsection{Risk Reasoning and Safety in Driving.}  
In the driving domain, numerous works have explored identifying risks that directly influence decision-making and overall safety~\cite{ma2024videotokensparsificationefficient, zhou2024hintspromptenhancingvisual, Wang2022PotentialRA, Aslantas2025EvaluatingTP}. Goal-oriented importance estimation in on-road videos~\cite{gao2019goalorientedobjectimportanceestimation} and deep spatio-temporal importance prediction~\cite{Understandingintent} focus on highlighting objects that affect vehicle navigation, while interaction-graph approaches~\cite{Zhang2020InteractionGF} explicitly model agent–agent relations to infer critical elements for future behavior. Advancing toward risk-aware reasoning, causal inference methods have been introduced to identify objects influencing driver actions, such as those that lead to stops~\cite{Li2020WhoMD}, and adaptive situational awareness systems~\cite{Wu2022TowardAA} aim to prioritize scene elements in complex urban contexts. More recently, efforts such as joint risk localization and captioning~\cite{malla2023drama} attempt to both detect hazardous objects and generate textual explanations of their impact on the ego-vehicle, while others explore risk assessment based on driving behaviors that violate safety commonsense~\cite{10933959}.% \textcolor{blue}{\textbf{Don't see relevancy of this for our work. Either we talk about other risk or safety related works in this domain or say no works in this direction have worked, and there is a need:}
% }

\subsection{MLLMs in Autonomous Driving.} 

MLLMs have recently garnered significant interest for their ability to analyze non-textual modalities, such as images and point clouds, through language-based reasoning~\cite{sima2025drivelmdrivinggraphvisual, zhou2024embodiedunderstandingdrivingscenarios, mao2023gpt, ding2024holisticautonomousdrivingunderstanding}. Leveraging their flexibility, MLLMs have been applied to various risk-related driving tasks. For instance, HiLM-D~\cite{ding2025hilmdenhancingmllmsmultiscale} utilizes hybrid-resolution perception to jointly detect safety-critical objects and predict their intentions, thereby improving risk localization. Likewise, MLLM-SUL~\cite{fan2024mllmsulmultimodallargelanguage} introduces a dual-branch visual encoding framework integrated with LLaMA-based reasoning to perform semantic risk inference and hazardous agent localization, building upon DRAMA-ROLISP~\cite{ding2025hilmdenhancingmllmsmultiscale}. Extending beyond single-vehicle contexts, V2V-LLM~\cite{chiu2025v2vllm} advances risk reasoning into cooperative multi-vehicle environments via shared perceptual fusion, enabling collaborative assessment across multiple agents. 

However, despite these advancements, no existing work has explicitly addressed safety as an integral objective, leaving a crucial gap in bridging risk understanding with actionable safety guidance. Motivated by this gap, our work seeks to integrate both risk reasoning and safety assessment within the MLLM paradigm.

\begin{figure}[t]
    \centering
    \begin{tcolorbox}[colback=gray!10, colframe=black, boxrule=0pt]
    \textbf{System:} You are an expert driving scene summarizer. Analyze multimodal driving data and produce a concise, geometry-aware summary.

    \vspace{0.5em}
    \textbf{User:} Given the multimodal inputs for a driving video:
    
    \textbf{Spatial context ($\mathcal{S}_t$)}: \{example\}\\
    \textbf{Motion dynamics ($\mathcal{M}_t$)}: \{example\}\\
    \textbf{Depth context ($\mathcal{D}_t$}): \{example\}\\
    \textbf{Video description ($d_v$}): \{example\}
    Generate a structured natural language summary that includes:
    \begin{enumerate}
        \item Ego vehicle maneuvers (\textit{lane changes}, \textit{turns}, stability)
        \item Behavior of surrounding agents (\textit{approaching}, \textit{overtaking}, \textit{falling behind})
        \item Lane position and road context
        \item Bounding boxes, depth, and motion cues for key objects
    \end{enumerate}
    
    The output should be a clear paragraph suitable for downstream reasoning tasks.
    \end{tcolorbox}
    \caption{Prompt template $P$ used to guide the LLM in generating structured, 
    spatially grounded summaries of driving scenes.}
    \label{fig:prompt_template}
    \vspace{-.4cm}
\end{figure}

% \subsection{Evaluation Protocols.}  
% Traditional language metrics (BLEU, METEOR, CIDEr) \cite{coco-eval} are widely used in captioning but are ill-suited for evaluating driving instructions. Recent works highlight the need for task-specific metrics in safety-critical domains \cite{eval-safety}. We contribute a rule-based evaluation protocol grounded in risk keywords (Table~\ref{tab:rules}, Algorithm~\ref{alg:safety}) and report accuracy and F1 score to directly assess the correctness and reliability of generated safety suggestions.  

\begin{figure*}[t]
    \centering
    \includegraphics[width=1\linewidth]{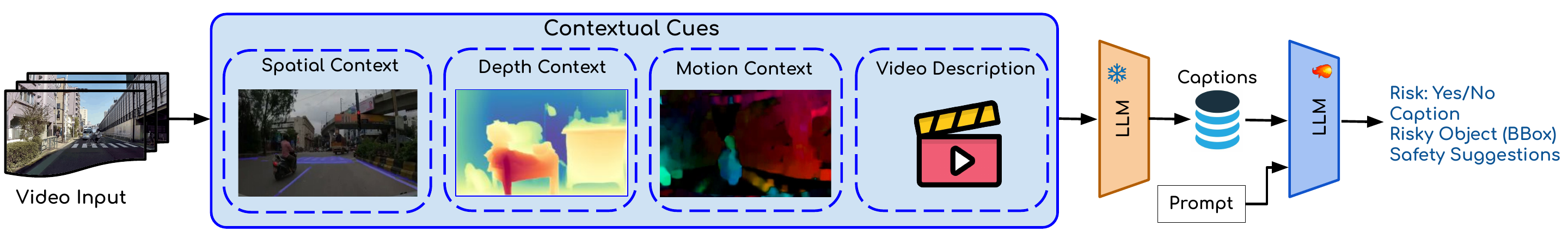}
    \caption{\textbf{Our proposed DriveSafe framework for the caption generation and safety suggestion task in driving. We first derive contextual cues to guide caption generation, and then use the resulting captions for risk assessment and safety suggestion.}}
    \label{fig:proposed_framework}
\vspace{-.2cm}
\end{figure*}

\section{DriveSafe Framework}

%We propose DriveSafe, which goes beyond conventional risk identification by jointly producing risk and suggesting corresponding safety actions. Unlike prior approaches that focus solely on risk prediction~\cite{malla2023drama, bolya2022token, zhou2024hintspromptenhancingvisual}, our method leverages multimodal context to provide safety actions. 
Given a video sequence $v$ our framework predicts a tuple: $(\hat{r}, C_r, b_r, s_r) = \mathcal{F}(v),$  where $\hat{r}$ denotes \textit{yes/no} for the risk, $C_r$ is the generated risk caption, $b_r$ is the bounding box of the visually grounded risky object, and $s_r$ is the associated safety suggestion for the risky behavior. The proposed framework is structured in two stages:
\begin{itemize}
    \item \textbf{Caption Generation} (\cref{subsec:cg}): Our framework generates scene-level captions enriched with multimodal contextual cues and geometric awareness of the driving environment.
    \item \textbf{Risk Assessment and Safety Suggestion} (\cref{subsec:riskassesment}): The generated captions are processed with an LLM to infer a binary risk classification $\hat{r}$, a risk caption $C_r$, risk-related keywords $\hat{K}$, and bounding boxes of risky objects $\hat{b}$. Safety suggestions are then derived from the predicted keywords $\hat{K}$. While the framework can operate in a zero-shot setting, we additionally fine-tune the LLM using lightweight adapters to enhance both performance and robustness.
\end{itemize}

\subsection{Caption Generation}
\label{subsec:cg}
Given a video sequence $v = \{f_1, f_2, \dots, f_T\}$, where each frame $f_t \in \mathbb{R}^{H \times W \times 3}$ and $T$ denotes the number of frames with spatial resolution $(H, W)$, we extract three types of contextual information. The spatial context $\mathcal{S}_t$ is obtained using road and lane segmentation masks that delineate boundaries and drivable areas. The motion context $\mathcal{M}_t$ is derived from optical flow, highlighting regions of relative movement corresponding to surrounding agents such as vehicles, cyclists, and pedestrians. Finally, the depth context $\mathcal{D}_t$ is computed at the object level, ensuring that each detected object in the surrounding (e.g., vehicles, pedestrians, traffic lights, and barriers) is associated with a depth estimate that reflects its spatial proximity to the ego-vehicle.  

While $\mathcal{S}_t$, $\mathcal{M}_t$, and $\mathcal{D}_t$ capture frame-level spatial, motion, and depth cues, we further enrich the representation with video-level semantics. Specifically, a high-level description $d_v$ is generated using a MLLM, summarizing the global scene context, such as ``\textit{a car is slowing down in front}'' or ``\textit{a pedestrian is crossing from the right}''. This global description complements the fine-grained frame-level cues by providing a holistic understanding of driving scenario.  

The complete multimodal representation of a video is defined as:
\begin{equation}
    X_v = \{\{\mathcal{S}_t, \mathcal{M}_t, \mathcal{D}_t\}_{t=1}^T, d_v\}.
\end{equation}
where frame-level spatial, motion, and depth contexts are complemented by the video-level semantic description. To leverage this representation, we construct a structured prompt $P(X_v)$ (illustrated in Fig.~\ref{fig:prompt_template}) that integrates both frame-level and global information into a unified input for the LLM $F_\theta$. The model then fuses these modalities to produce a geometry-aware description of the video:
\begin{equation}
    C_v = F_\theta(P(X_v)),
\end{equation}
where $C_v$ denotes the generated caption for the sequence.

\begin{table*}[htbp]
\centering

\label{tab:rules}
\begin{tabular}{p{0.25\textwidth} p{0.70\textwidth}}
\toprule
\textbf{Safety Suggestions} & \textbf{Risk-related Keywords} \\
\midrule
\textbf{(Must) Stop} & Pedestrian crossing (19); Stopped vehicle (860); Crosswalk (105); Traffic light red (751); Traffic light yellow (5); Traffic congestion (877) \\
\textbf{Be aware / cautious} (object may affect future but no direct influence) & Cyclist nearby (8); Pedestrian nearby (12); Traffic signal (19); Traffic sign (62); Leading vehicle (151) \\
\textbf{Slow down} & Slowing (277); Pedestrian ahead (4); Heavy traffic (159); Cut-in (8); Cyclist (12) \\
\textbf{Carefully maneuver} (around important object) & Parked vehicle (28); Traffic cones (9) \\
\textbf{Follow the vehicle ahead} & Vehicle in front (234); Following traffic (44); Same lane (228); Near the intersection (77) \\
\textbf{Yield} & Merging traffic (7); Vulnerable Road User (VRU) (37); Right of way (123); Oncoming traffic (109); At the crosswalk (86); \\
\textbf{Start moving} & Traffic cleared (18); Vehicle ahead moved (36); Traffic light green (10) \\
\textbf{NA} & Irrelevant (5); Background (5); No decision (5) \\
\bottomrule
\end{tabular}
\caption{Mapping between safety suggestions and their corresponding risk-related keywords in the DRAMA~\cite{malla2023drama} dataset. Instances of keywords are shown in parentheses.}
\label{tab:safety_risk_correspondance}
\vspace{-0.3cm} 
\end{table*}

% The captions $y_v$ generated in the previous stage serve as an interpretable intermediate representation for downstream risk reasoning. Rather than operating directly on raw visual features, we leverage the descriptive content of $y_v$ as input to an instruction-tuned LLM $f_\theta$, which is prompted to perform a sequence of risk-aware tasks.  

\begin{figure}[t]
  \centering
  \begin{adjustbox}{max width=\textwidth}
    \begin{tcolorbox}[colback=gray!10, colframe=black, boxrule=0pt]
    \small
    \textbf{System:} You are a risk-aware driving scene analyst. Given a caption $C_v$ with object tags and bounding boxes, 
    analyze risks, explain them, and suggest safety-aware actions.

    \vspace{0.4em}
    \textbf{Inputs:}
    Caption ($C_v$) \\
    \textbf{Outputs (per object):}
    \begin{enumerate}
      \item Risk label $\hat{r} \in \{Yes, No\}$
      \item Risk caption ($C_r$)
      \item Risk-related keywords ($\hat{K}$)
      \item Object Localization - Bounding box ($\hat{b}$)
    \end{enumerate}

    Reasoning should be accurate, structured, and safety-focused.

    \vspace{0.4em}
    \textbf{Example Input $C_v$:} 
    ``A cyclist [bbox: 612, 350, 720, 480] is crossing from the left; a red car [bbox: 1000, 400, 1200, 550] is stopped in the ego lane."

    \textbf{Example Output:} \\
    1) $\hat{r}=$Yes; $C_r$: cyclist crossing may intersect ego path; $\hat{K}=$\{Cyclist, Crossing\}; $\hat{b}=$[612, 350, 720, 480].\\
    2) $\hat{r}=$Yes; $C_r$: red car stopped in ego lane blocks motion; $\hat{K}=$\{Stopped vehicle\}; $\hat{b}=$[1000, 400, 1200, 550].
    \end{tcolorbox}
  \end{adjustbox}
  \caption{Risk-aware prompt template consistent with notation. 
  The LLM $F_\theta$ maps $C_v$ to $(\hat{r}, C_r, \hat{K}, \hat{b})$.}
  \label{fig:risk_prompt_template}
  \vspace{-.2cm}
\end{figure}

\subsection{Risk Assessment and Safety Suggestion}
\label{subsec:riskassesment}
\noindent\textbf{Zero-Shot:} Given the geometric-aware caption  $C_v$, we prompt the LLM $F_\theta$ using a structured template as illustrated in Fig.~\ref{fig:risk_prompt_template} to produce risk-aware outputs. Specifically, the model generates: 
\begin{itemize}
    \item \textit{Risk label:} a binary indicator of whether the scene involves risk,
    \item \textit{Refined risk caption:} a descriptive explanation specifying the risky object and its spatial location,
    \item \textit{Risk-associated keywords:} salient terms that capture the risky behavior or objects involved, and
    \item \textit{Bounding box:} the localized coordinates of the identified risky object.
\end{itemize}   

Formally, this process is defined as:
\begin{equation}
    (\hat{r}, \hat{C_r}, \hat{K}, \hat{b}) = f_\theta(C_v),
\end{equation}
where $\hat{r} \in \{Yes,No\}$ denotes the binary risk classification, $\hat{C_r}$ is the generated risk caption, $\hat{K}$ is the set of extracted risk-related keywords, and $\hat{b} = (\hat{x}_{\min}, \hat{y}_{\min}, \hat{x}_{\max}, \hat{y}_{\max})$ represents the predicted bounding box of the risky object. 

To derive \textit{safety suggestions}, the extracted keywords $\hat{K}$ are mapped to corresponding driving action categories using a predefined rule set (see Table~\ref{tab:safety_risk_correspondance}). The final safety suggestion $s_r$ is then obtained through keyword matching with ground-truth categories
\begin{equation}
    s_r = g(\hat{K}),
\end{equation}
where $g(\cdot)$ denotes the mapping function from risk keywords to safety suggestions.

% \subsection{Risk Assessment and Safety Suggestion (Adapter-based Finetuning)}

\noindent\textbf{Adapter Finetuning:} In the fine-tuning stage, we adapt the LLM $F_{\theta}$ using lightweight parameter-efficient adapters trained on a dataset of caption–instruction–response triplets. Given an input caption $c_v$ and instruction $q$, the model is supervised to generate a structured response $\hat{a} = (\hat{r}, \hat{C_r}, \hat{K}, \hat{\mathbf{b}})$. Formally, the output is defined as:
\begin{equation}
    \hat{a} = F_{\theta}(C_v, q),
\end{equation}
where $\hat{a}$ represents the predicted risk reasoning sequence. 

The training objective minimizing the negative log-likelihood of the ground-truth structured response:
\begin{equation}
    \mathcal{L} = - \sum_{t=1}^{T} \log p_{\theta}(a_t \mid C_v, q, a_{<t}),
\end{equation}
where $a_t$ is the $t^{\text{th}}$ token of the response and $a_{<t}$ are the previously generated tokens.  

This adapter-based fine-tuning explicitly aligns video captions with risk classification $\hat{r}$, descriptive risk captions $C_r$, risk-related keywords $\hat{K}$, and bounding boxes $\hat{\mathbf{b}}$. Safety suggestions are then derived directly from the generated risk keywords $\hat{K}$ via string matching to predefined categories. Model performance is evaluated using accuracy and $F_1$ scores over nine safety suggestion classes (see Table~\ref{tab:safety_risk_correspondance}), thereby providing a direct measure of both correctness and reliability.

\begin{table}[htbp]
    \centering
    \setlength{\tabcolsep}{2pt} % Adjust column spacing
    \renewcommand{\arraystretch}{1.15} % Row height
    \scriptsize % shrink whole table numbers
    \begin{tabular}{>{\footnotesize\raggedright\arraybackslash}p{2.25cm}ccccccccc}
        \toprule
        \textbf{Method} & \textbf{B1$\uparrow$} & \textbf{B4$\uparrow$} & \textbf{M$\uparrow$} & \textbf{R$\uparrow$} & \textbf{C$\uparrow$} & \textbf{S$\uparrow$} & \textbf{CLAIR} & \textbf{\makecell{Mean\\IoU}} & \textbf{\makecell{Acc@\\0.5}} \\
        \midrule
        \rowcolor{blue!10} LCP~\cite{malla2023drama} & 73.9 & 54.7 & 39.1 & 70.0 & \textbf{3.7} & 56.0 & -- & 61.4 & 68.4\\
        \rowcolor{blue!10} VTS~\cite{ma2024videotokensparsificationefficient} & 75.3 & 55.8 & 40.7 & 74.7 & 2.8 & 58.0 & -- & 66.8 & \underline{74.4} \\
        \rowcolor{blue!10} LLaVA-v1.5$^\dagger$~\cite{liu2024improved} & 75.8 & 56.1 & 41.0 & 78.0 & 2.9 & 58.4 & -- & -- & -- \\
        \rowcolor{blue!10} Efficient HoP~\cite{zhou2024hintspromptenhancingvisual} & \underline{76.0} & 56.2 & 41.3 & 78.5 & 2.7 & 58.8 & -- & -- & -- \\
        \rowcolor{blue!10} HoP~\cite{zhou2024hintspromptenhancingvisual} & \textbf{76.2} & \underline{56.3} & \underline{41.7} & \underline{79.8} & 2.87 & \underline{59.1} & -- & -- & -- \\ 
        \midrule
        \rowcolor{green!15} Qwen2.5-VL~\cite{bai2025qwen25vltechnicalreport} & 27.72 & 4.94 & 18.72 & 26.15 & 0.08 & 16.30 & 24.33 & 0.0 & 0.0 \\
        \rowcolor{green!15} LLaVA-NeXT~\cite{li2024llavanextinterleavetacklingmultiimagevideo} & 28.17 & 4.30 & 18.87 & 26.60 & 0.06 & 16.76 & 23.22 & \textbf{83.6} & 13.1 \\
        \rowcolor{green!15} VideoLLaMA3~\cite{zhang2025videollama3frontiermultimodal} & 26.73 & 3.24 & 21.36 & 23.70 & 0.07 & 11.91 & 24.55 & 3.9 & 4.3 \\
        \midrule
        \rowcolor{orange!15} \textbf{DriveSafe-Zeroshot} & 30.65 & 10.55 & 33.92 & 35.70 & 0.12 & 22.11 & \underline{30.47} & \underline{82.4} & 0.9 \\
        \rowcolor{orange!15} \textbf{DriveSafe-Finetuned} & 64.47 & \textbf{60.38} & \textbf{64.78} & \textbf{80.85} & \underline{3.27} & \textbf{64.91} & \textbf{58.93} & 59.8 & \textbf{74.8} \\
        \bottomrule
    \end{tabular}
    \caption{Performance comparison of \textbf{Caption Generation} and \textbf{Risky Object Grounding} across \colorbox{blue!15}{Existing Methods}, \colorbox{green!15}{General VLMs}, and \colorbox{orange!15}{DriveSafe} on the DRAMA~\cite{malla2023drama} dataset.}
    \label{tab:captioning_results}
\vspace{-.1cm}
\end{table}

%Captioning metrics (B1: BLEU-1~\cite{papineni2002bleu}, B4: BLEU-4~\cite{papineni2002bleu}, M: METEOR~\cite{banerjee2005meteor}, R: ROUGE~\cite{lin2004rouge}, C: CIDEr~\cite{vedantam2015cider}, S: SPICE~\cite{anderson2016spice}, CLAIR~\cite{chan2023clairevaluatingimagecaptions}) are reported alongside object grounding performance (Mean IoU and Accuracy for IoU $>$ 0.5).

\begin{table}[h]
    \label{tab:saftey_eval}
    \centering
    \begin{tabular}{lcc}
        \toprule
        \textbf{Model} & \textbf{Accuracy} & \textbf{$F_1$ (Weighted)} \\
        \midrule
        LLaVA-Next~\cite{li2024llavanextinterleavetacklingmultiimagevideo}
        & 15.83 & 20.95 \\
        VideoLLaMA 3~\cite{zhang2025videollama3frontiermultimodal} & 13.00 & 19.55  \\
        Qwen2.5 VL~\cite{bai2025qwen25vltechnicalreport} & 18.88 & 23.19  \\
        \midrule
        DriveSafe - Zeroshot & \underline{23.49} & \underline{24.80} \\
        DriveSafe - Finetuned & \textbf{52.85} & \textbf{37.15} \\
        \bottomrule
    \end{tabular}
    \caption{Performance comparison of \textbf{Safety Suggestion} prediction across General-VLMs and DriveSafe on the DRAMA~\cite{malla2023drama}.}
    \label{tab:model_comparison}
    \vspace{-0.4cm} 
\end{table}

%%%%%%%%%%%%%%%%%%%%%%%%%%%%%%%%
\section{Experiments}
\label{benchmark-sec}

\begin{figure}[t]
    \centering
    \includegraphics[width=1\linewidth]{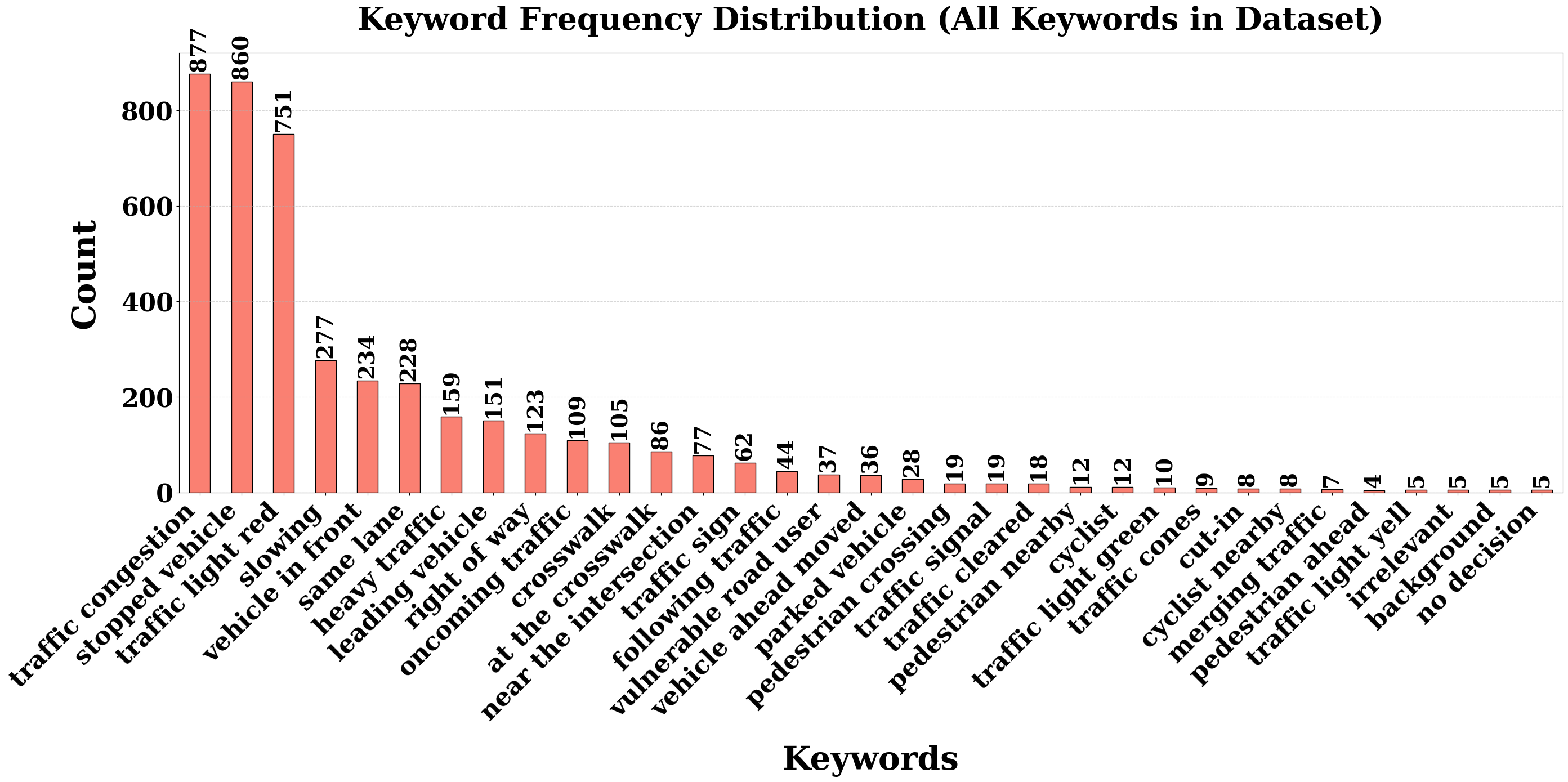}
    \caption{Distribution of driving decision categories in the curated test set. 
    Each bar corresponds to a safety suggestion category, showing the aggregated frequency of its representative keywords.}
    \label{fig:category_distribution}
    \vspace{-.2cm}
\end{figure}

\subsection{Dataset}
\label{data-capture}

We use DRAMA dataset~\cite{malla2023drama} which consists of 17K interactive driving scenario videos with rich annotations, including driving risk-related captions, nine categories of safety suggestions, and bounding boxes for important objects to enable visual grounding. 
To evaluate safety suggestions in the context of risky behaviors, we re-organize the DRAMA dataset by explicitly linking safety suggestions with the corresponding risk-related keywords derived from risky behaviors described in the captions, as illustrated in Table~\ref{tab:safety_risk_correspondance}.
The test set of DRAMA dataset is manually reviewed by annotators, and mapped the nine predefined safety suggestions one-to-one with risk-related keywords by carefully observing videos and their captions. This re-annotation enables a more reliable safety evaluation. 
The distribution of safety-critical cues and their frequency of occurrence are shown in Fig.~\ref{fig:category_distribution}. At the dataset level, the five most frequent risk-related keywords were \textit{traffic congestion}, \textit{stopped vehicle}, \textit{traffic light red}, \textit{slowing}, and \textit{vehicle in front}.
%However, autoregressively generating safety suggestions based only on captions does not guarantee proper conditioning on the risky behaviors described in the scene. However, we highlight that reliable safety evaluation also requires explicit connections between risky behaviors and corresponding driver actions. We annotate these in the form of keywords.

%For annotation, the test set of DRAMA dataset The To minimize annotation overhead, only the test set is enriched with these risk tags. The process involves manually reviewing videos in the test split to identify a consistent set of risk-related terms, aligned with the nine predefined safety suggestion categories defined by DRAMA.

%Risk-behavior keywords were allocated to the nine predefined safety suggestion categories: (must) Stop (6 keywords), Be aware or cautious (5), Slow down (6), Carefully maneuver (2), Follow the vehicle ahead (4), Yield (6), Start moving (3), and N/A (3). Across the dataset, this resulted in 4,377 keyword instances spanning 33 unique terms. Within each category, a dominant keyword emerged, such as traffic congestion (877 instances) for (must) Stop, leading vehicle (151) for Be aware or cautious, slowing (277) for Slow down, parked vehicle (28) for Carefully maneuver, vehicle in front (234) for Follow the vehicle ahead, right of way (123) for Yield, vehicle ahead moved (36) for Start moving, and irrelevant (1) for N/A. At the dataset level, the five most frequent keywords were traffic congestion (877), stopped vehicle (860), traffic light red (751), slowing (277), and vehicle in front (234). 

\begin{figure*}[t]
    \centering
    \includegraphics[width=1\linewidth]{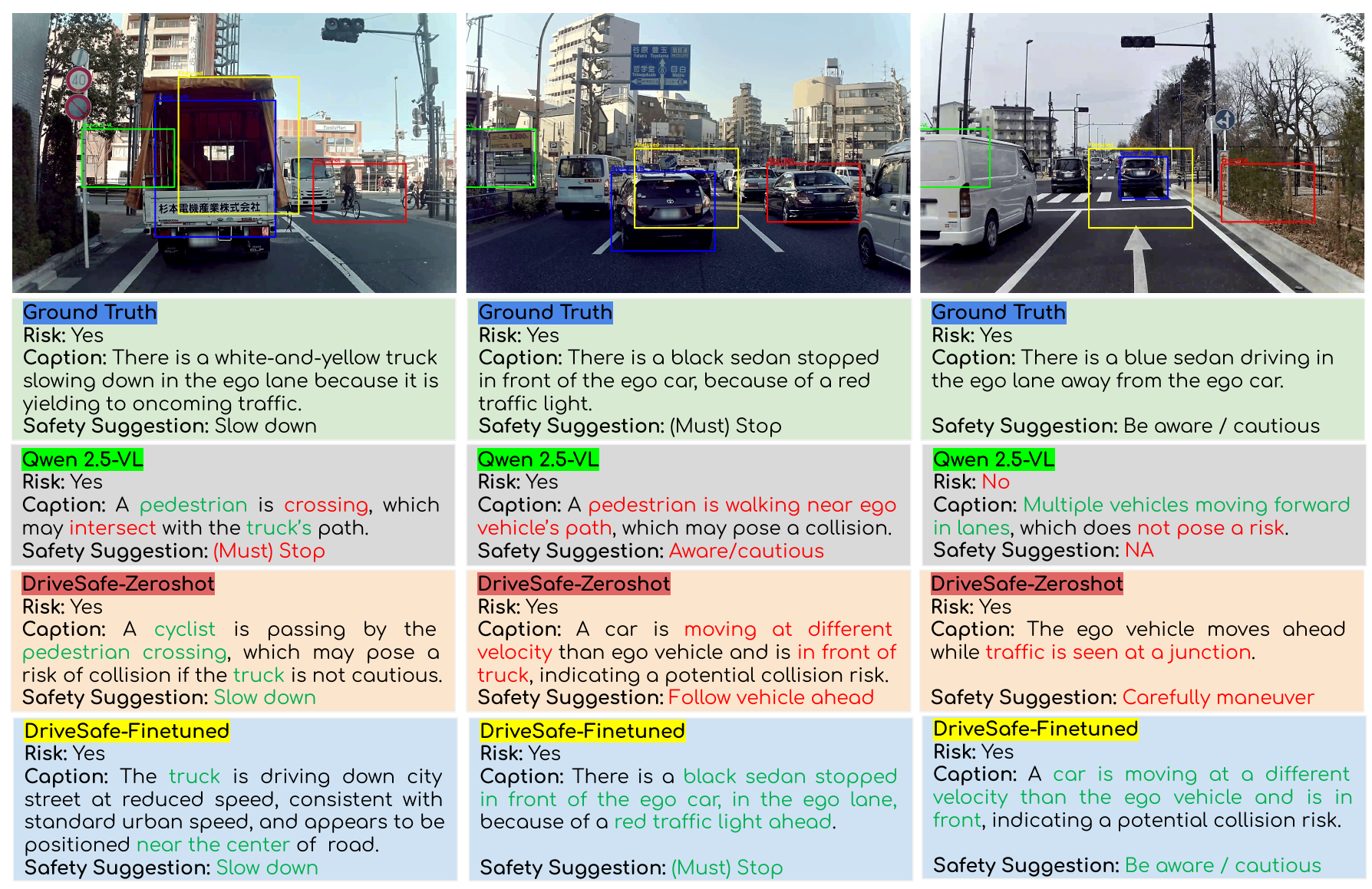}
    \caption{Qualitative comparison of DriveSafe-ZeroShot, DriveSafe-Finetuned, and Qwen2.5-VL~\cite{bai2025qwen25vltechnicalreport} on three driving scenarios from the DRAMA dataset~\cite{malla2023drama}. Risky object grounding is shown with bounding boxes so is respective models with text highlighting, while generated captions and safety suggestions are marked as correct (\textcolor{good}{green}) or incorrect (\textcolor{bad}{red}).}
\label{fig:qual_results}
\vspace{-.2cm}
\end{figure*}

\subsection{Experimental Settings}

\noindent\textbf{Models.}
We compare our framework with several video-based MLLMs. We consider risk assessment models such as LCP~\cite{malla2023drama}, VTS~\cite{bolya2022token}, LLaVA-v1.5~\cite{liu2024improved} and HoP~\cite{zhou2024hintspromptenhancingvisual}. In addition, we also compare with several open source MLLMs such as Qwen2.5~\cite{bai2025qwen25vltechnicalreport}, LLaVA-NeXT~\cite{li2024llavanextinterleavetacklingmultiimagevideo}, and VideoLLaMA  3~\cite{zhang2025videollama3frontiermultimodal} to assess their performance on risk prediction, risky object localization and safety suggestion prediction. 

\noindent\textbf{Metrics.}
We evaluate caption generation using standard metrics, including BLEU-1 (B1)~\cite{papineni2002bleu}, BLEU-4 (B4)~\cite{papineni2002bleu}, METEOR (M)~\cite{banerjee2005meteor}, ROUGE-L (R)~\cite{lin2004rouge}, CIDEr (C)~\cite{vedantam2015cider}, and SPICE (S)~\cite{anderson2016spice}. In addition, we incorporate CLAIR~\cite{chan2023clairevaluatingimagecaptions}, a recently proposed LLM-based evaluation metric designed to better capture semantic quality beyond n-gram overlap. For localization, we report Intersection-over-Union (IoU) as the evaluation metric.

Safety suggestions are generated using the ground-truth rule map that links risk-related keywords to suggestion categories (\cref{tab:rules}). Following DRAMA~\cite{malla2023drama}, we consider eight safety suggestion classes (excluding NA). Accuracy and $F_1$ score are reported as the primary metrics for evaluating suggestion quality. 

\noindent\textbf{Implementation Details.} For spatial context, we use HybridNets~\cite{vu2022hybridnetsendtoendperceptionnetwork}, motion cues are obtained by computing dense optical flow using the Farnebäck algorithm from OpenCV, and depth cues are estimated using DepthAnything-v2~\cite{yang2024depthv2}. For fine-tuning, we adopt the LLaMA-Adapter framework~\cite{zhang2024llamaadapter}. Training is performed with a batch size of 4, learning rate of $2\times10^{-5}$, and weight decay of 0.01 for 5 epochs on NVIDIA A6000 (40GB) GPU. To improve efficiency, we apply gradient checkpointing and mixed precision, along with a linear warmup schedule over the first 10\% of training steps. For zero-shot inference, we employ LLaMA-3.1-8B~\cite{grattafiori2024llama3herdmodels}, which directly conditions on captions without additional fine-tuning.

\subsection{Quantitative results}

\noindent\textbf{Caption Generation.} 
Table~\ref{tab:captioning_results} presents the quantitative comparison across captioning metrics (BLEU~\cite{papineni2002bleu}, METEOR~\cite{banerjee2005meteor}, ROUGE~\cite{lin2004rouge}, CIDEr~\cite{vedantam2015cider}, SPICE~\cite{anderson2016spice} and CLAIR~\cite{chan2023clairevaluatingimagecaptions}). DriveSafe-Finetuned consistently outperforms its zero-shot counterpart across all metrics, with BLEU-4 improving from 10.55 to 60.38, METEOR from 33.92 to 64.78, and ROUGE from 35.70 to 80.85. These gains highlight the effectiveness of our caption-to-risk assessment adapter-based instruction tuning in capturing safety-critical driving narratives. 

Compared to specialized driving models such as HoP~\cite{zhou2024hintspromptenhancingvisual} (BLEU-4: 56.3, METEOR: 41.7) and LLaVA-v1.5~\cite{liu2024improved} (BLEU-4: 56.1, METEOR: 41.0), DriveSafe-Finetuned achieves superior performance despite relying only on lightweight adapter tuning rather than full-scale video pretraining. In contrast, general-purpose VLMs show severe degradation on driving scenarios, with Qwen2.5-VL~\cite{bai2025qwen25vltechnicalreport}, LLaVA-NeXT~\cite{li2024llavanextinterleavetacklingmultiimagevideo}, and VideoLLaMA-3~\cite{zhang2025videollama3frontiermultimodal} achieving BLEU-4 scores below 30 and METEOR scores below 22. This clearly underscores the challenges of directly adapting general VLMs to safety-critical domains, which our domain-specific finetuning strategy effectively bridges this gap.
% Table~\ref{tab:captioning_results} presents the quantitative comparison across captioning metrics (BLEU~\cite{papineni2002bleu}, METEOR~\cite{banerjee2005meteor}, ROUGE~\cite{lin2004rouge}, CIDEr~\cite{vedantam2015cider}, SPICE~\cite{anderson2016spice} and CLAIR~\cite{chan2023clairevaluatingimagecaptions}). Our DriveSafe-Finetuned significantly outperforms its zero-shot variant, improving BLEU-4~\cite{papineni2002bleu} from 10.55 to 60.38, METEOR~\cite{banerjee2005meteor} from 33.92 to 64.78, and ROUGE~\cite{lin2004rouge}
% from 35.70 to 80.85. These results highlight the effectiveness of our caption-to-risk assessment adapter-based instruction tuning in aligning model outputs with safety-critical driving narratives.

% When compared to prior video-based baselines such as LLaVA-v1.5~\cite{liu2024improved} and HoP~\cite{zhou2024hintspromptenhancingvisual}, DriveSafe-Finetuned achieves competitive performance across standard captioning metrics, substantially narrowing the gap despite relying on structured prompts and lightweight adapter tuning rather than full-scale video pretraining. Moreover, unlike general-purpose VLMs (Qwen2.5-VL~\cite{bai2025qwen25vltechnicalreport}, LLaVA-NeXT~\cite{li2024llavanextinterleavetacklingmultiimagevideo}, VideoLLaMA-3~\cite{zhang2025videollama3frontiermultimodal}), which underperform on driving-specific captioning tasks, DriveSafe-Zeroshot demonstrates strong semantic alignment with safety-critical contexts.
\noindent\textbf{Risky Object Grounding.} For object grounding, we evaluate performance using MeanIoU and Acc@0.5. General-purpose VLMs perform poorly in this setting, with Qwen2.5-VL~\cite{bai2025qwen25vltechnicalreport} and VideoLLaMA-3~\cite{zhang2025videollama3frontiermultimodal} nearly fail, while LLaVA-NeXT~\cite{li2024llavanextinterleavetacklingmultiimagevideo} achieves only limited accuracy (MeanIoU 83.6, Acc@0.5 13.1).
In contrast, DriveSafe-Finetuned shows strong localization ability, reaching an Acc@0.5 of 74.8, far surpassing all other models. This demonstrates its effectiveness in linking descriptive captions with precise spatial grounding, an ability that is essential for downstream safety reasoning. 

To further analyze grounding reliability, we evaluated accuracy and IoU across multiple thresholds. In the zero-shot setting, accuracy remains near-zero (e.g., 3.4\% at 0.1, dropping to 1.0\% at 0.4), while IoU appears deceptively high (rising from 35.8\% at 0.1 to 80.2\% at 0.4). This discrepancy arises from a few rare correct matches that inflate IoU scores, despite the model failing to capture most risky agents. In contrast, the finetuned model achieves both high accuracy and strong IoU jointly, i.e., 91.0\% / 90.0\% at 0.1, 83.0\% / 81.0\% at 0.3, and 79.0\% / 66.0\% at 0.4. 

Overall, these results highlight that IoU alone is an unreliable measure for safety-critical grounding, as zero-shot models can achieve seemingly high IoU while missing most risks. The consistent combination of high accuracy and IoU achieved by DriveSafe-Finetuned underscores the necessity of task-specific adaptation for reliable safety reasoning.

% At lower thresholds, the zero-shot model fails to provide meaningful grounding: for example, at 0.1 threshold, accuracy is only 3.4\%, while Mean IoU is 35.8\%. As the threshold increases, IoU appears to improve (e.g., 60.4\% at 0.2, 72.3\% at 0.3, and 80.2\% at 0.4), but this is an artifact of the extremely sparse correct predictions (accuracy collapsing to 1.6\%–1.0\%). In other words, the few surviving detections happen to align well spatially, inflating IoU without reflecting true performance.

% By contrast, the finetuned DriveSafe model achieves both high accuracy and strong IoU across thresholds. At 0.1, it delivers 91.0\% accuracy and 90.0\% IoU, and remains robust even under stricter thresholds (e.g., 84.0\% / 89.0\% at 0.2, 83.0\% / 81.0\% at 0.3, and 79.0\% / 76.0\% at 0.4). This joint improvement demonstrates that finetuning not only increases detection frequency but also ensures precise grounding of risky agents.

% Taken together, these results show that IoU alone is not a reliable measure of safety-critical grounding, since zero-shot models can show inflated IoU despite missing most risks. The combination of high accuracy and IoU achieved by finetuned DriveSafe highlights the necessity of task-specific adaptation for reliable safety reasoning.

\noindent \textbf{Safety Suggestion.} Table~\ref{tab:model_comparison} presents the accuracy and weighted $F_1$ scores for the safety suggestion task. General-purpose VLMs such as LLaVA-Next, VideoLLaMA-3, and Qwen2.5-VL perform poorly, with accuracies in the range of 13–19\% and weighted $F_1$ scores below 24\%. These results indicate that open-domain VLMs lack the fine-grained risk understanding necessary for safety-critical driving scenarios.
In contrast, our proposed DriveSafe model shows clear improvements. In the zeroshot setting, DriveSafe already surpasses all general VLM baselines, achieving 23.49\% accuracy and a 24.80 weighted $F_1$, demonstrating the benefit of structured multimodal prompting even without finetuning. After finetuning, DriveSafe achieves 52.85\% accuracy and 37.15 weighted $F_1$, more than doubling its zeroshot performance and substantially outperforming all baselines. On an NVIDIA A6000 GPU, LLaMA-Adapter 3.1 (8B) achieves a per-token latency of approximately 7--11 ms ($\sim$90--140 tokens/s), demonstrating suitability for near real-time deployment.

\subsection{Qualitative Analysis}

As shown in Fig.~\ref{fig:qual_results}, we observe in the first column that Qwen2.5-VL~\cite{bai2025qwen25vltechnicalreport} misidentifies risks, for example describing ``\textit{a pedestrian is crossing}" instead of a cyclist, shifting attention away from the true hazard. DriveSafe-ZeroShot also adds false context, e.g., ``\textit{a cyclist is passing by the pedestrian crossing}". In contrast, DriveSafe-Finetuned provides grounded interpretations such as ``\textit{the truck is driving at reduced speed on a city street}" with the correct suggestion ``\textit{Slow down}". In the second column, DriveSafe-ZeroShot exaggerates motion risk (e.g., ``\textit{a car moving at a different velocity in front of the truck}"), whereas DriveSafe-Finetuned correctly identifies the ``\textit{black sedan stopped in the ego lane}". In the third column, Qwen2.5-VL misses the critical ego-lane sedan, while DriveSafe-ZeroShot gives only a vague suggestion. DriveSafe-Finetuned instead highlights the collision risk and provides the precise recommendation ``\textit{Be aware / cautious}". These examples show that fine-tuning reduces hallucinations and exaggeration while improving focus on safety-critical evidence.

%Qwen2.5-VL~\cite{bai2025qwen25vltechnicalreport} misidentifies the scene, stating \textit{“A pedestrian is crossing”}—an incorrect classification of cyclist that shifts focus to the wrong risk context for the ego vehicle. Similarly, DriveSafe-Zeroshot introduces false context with \textit{“A cyclist is passing by the pedestrian crossing”}. In contrast, DriveSafe-Finetuned provides a grounded interpretation, noting \textit{“The truck is driving down a city street at reduced speed”} and giving the appropriate safety suggestion \textit{“Slow down”} (first column).
%In the second column DriveSafe-Zeroshot improves relevance but still misinterprets motion, such as labeling a car as \textit{``moving at a different velocity, in front of the truck''}, exaggerating risk. 
%In contrast, DriveSafe-Finetuned remains closely aligned with the scene: it captures the \textit{``black sedan stopped, in the ego lane''}. These examples show that fine-tuning reduces hallucinations and exaggeration, while enhancing focus on safety-critical evidences. In the third column, Qwen2.5-VL fails to capture the risk, stating \textit{“Multiple vehicles are moving forward”} does not pose an immediate risk, when in fact the ego-lane sedan is critical. DriveSafe-Zeroshot offers only a vague description, suggesting \textit{“Carefully maneuver”}. DriveSafe-Finetuned provides the most accurate interpretation: \textit{“A car is moving at a different velocity than the ego vehicle and is in front, indicating a potential collision risk,”} followed by the precise recommendation “Be aware / cautious.”

\tcbset{mybox/.style={
    colback=white,       % background = white (colorless)
    colframe=black,      % border color
    boxrule=0.5pt,       % border thickness
    sharp corners,       % square corners
    fonttitle=\bfseries, % bold titles if needed
    coltitle=black,      % title color
    left=2pt, right=2pt, top=2pt, bottom=2pt % padding
}}
\renewcommand{\resultbox}[2]{%
  \begin{tcolorbox}[colback=white, colframe=black, boxrule=0.5pt,
                    sharp corners, left=2pt, right=2pt, top=2pt, bottom=2pt]
    \textbf{#1:} #2
  \end{tcolorbox}%
}
\begin{figure}[h]
    \centering
    \begin{tabular}{c c}
        % ---------- First image ----------
        \begin{minipage}{0.45\linewidth}
            \centering
            {\scriptsize \textbf{Finetuned:} Be aware/cautious} \\[0.2em]
            \begin{tabular}{@{}c@{}}
                \rotatebox{90}{\parbox{2cm}{\centering\tiny \textbf{GT:} Be aware/cautious}} 
                \includegraphics[width=0.95\linewidth]{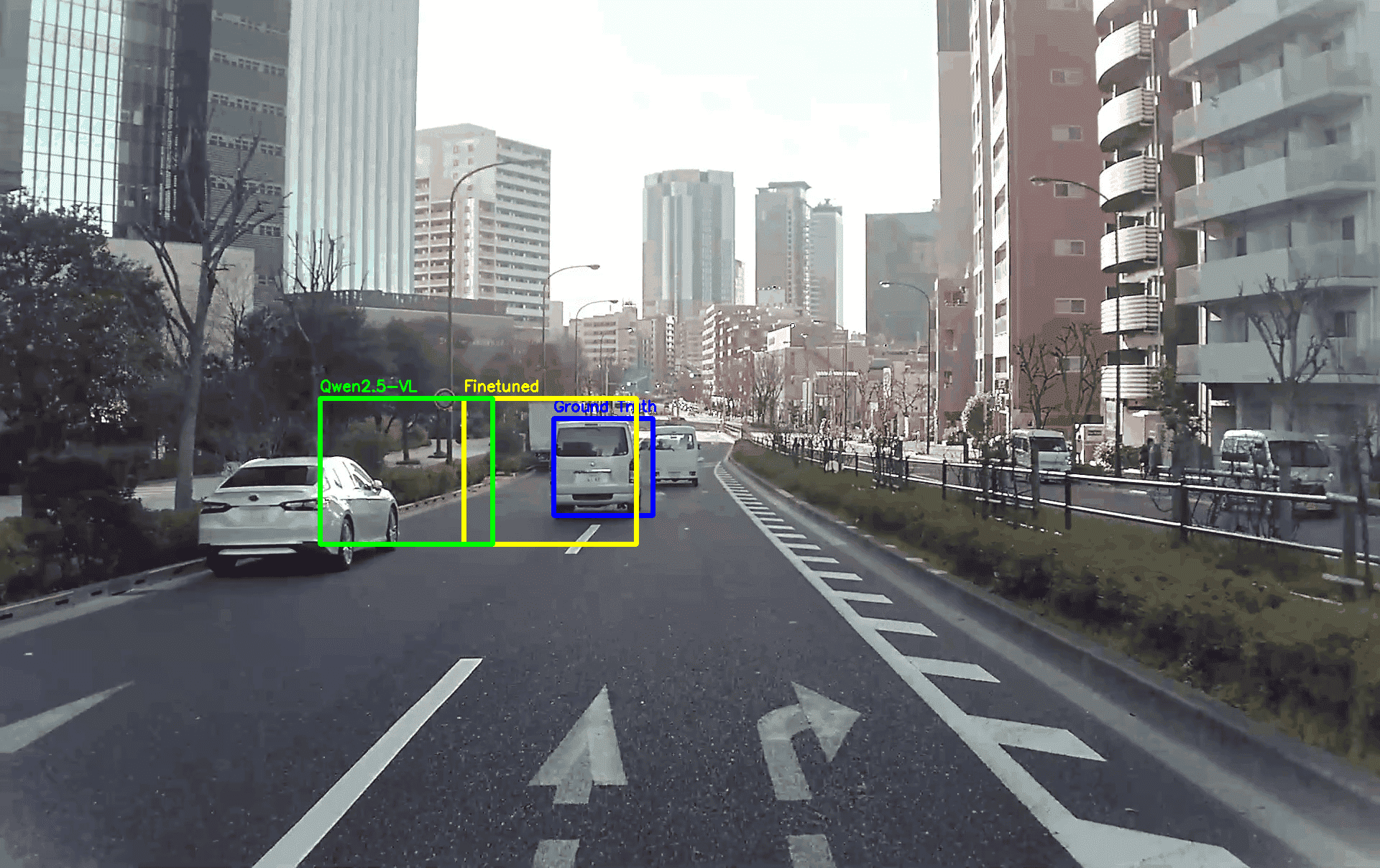}
            \end{tabular}
        \end{minipage} &
        % ---------- Second image ----------
        \begin{minipage}{0.45\linewidth}
            \centering
            {\scriptsize \textbf{Finetuned:} Slow down} \\[0.2em]
            \begin{tabular}{@{}c@{}}
                \rotatebox{90}{\parbox{2cm}{\centering\scriptsize \textbf{GT:} Slow down}} 
                \includegraphics[width=0.95\linewidth]{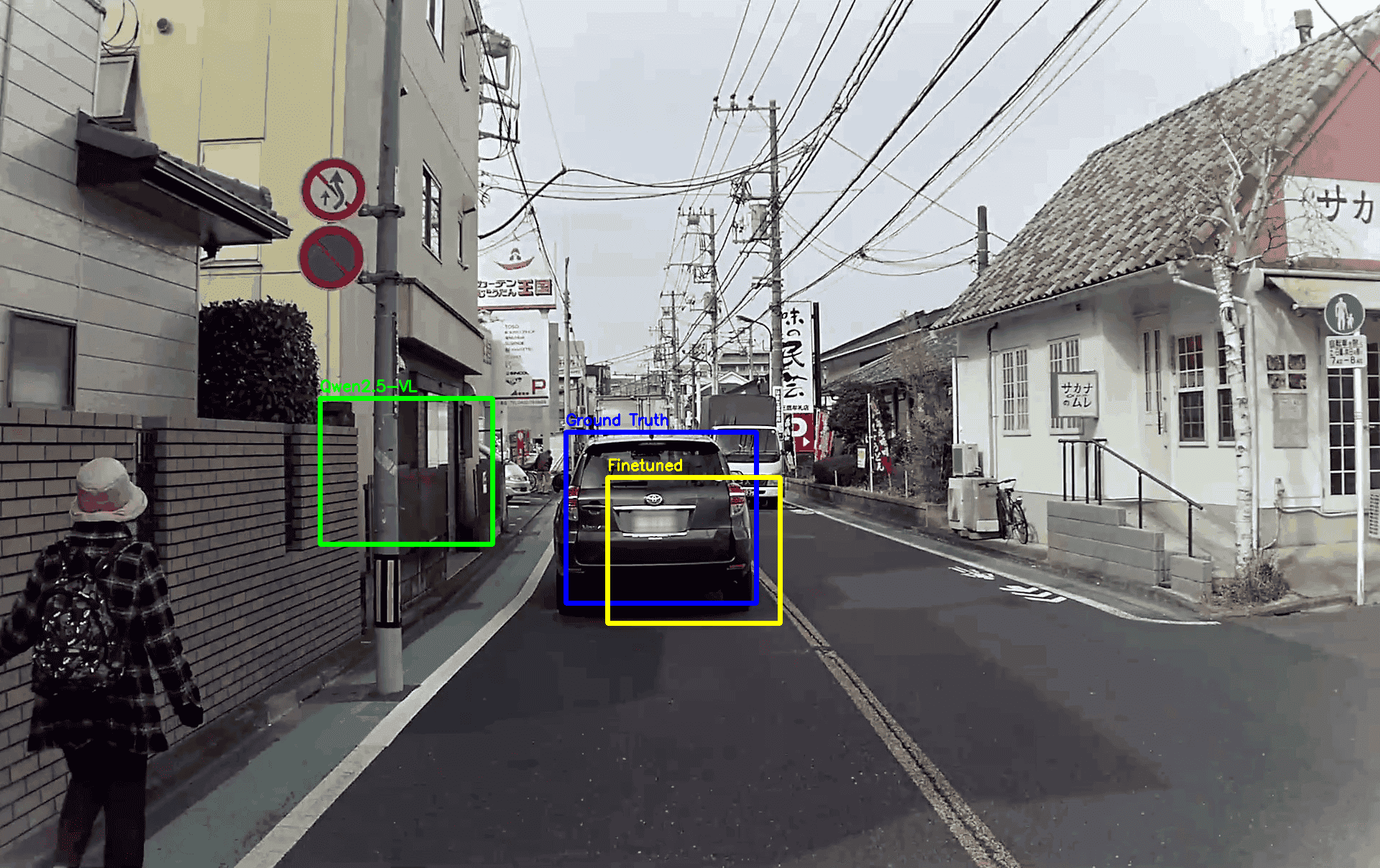}
            \end{tabular}
        \end{minipage} \\
        % ---------- Third image ----------
        \begin{minipage}{0.45\linewidth}
            \centering
            {\scriptsize \textbf{Finetuned:} Slow down} \\[0.2em]
            \begin{tabular}{@{}c@{}}
                \rotatebox{90}{\parbox{2cm}{\centering\scriptsize \textbf{GT:} Slow down}} 
                \includegraphics[width=0.95\linewidth]{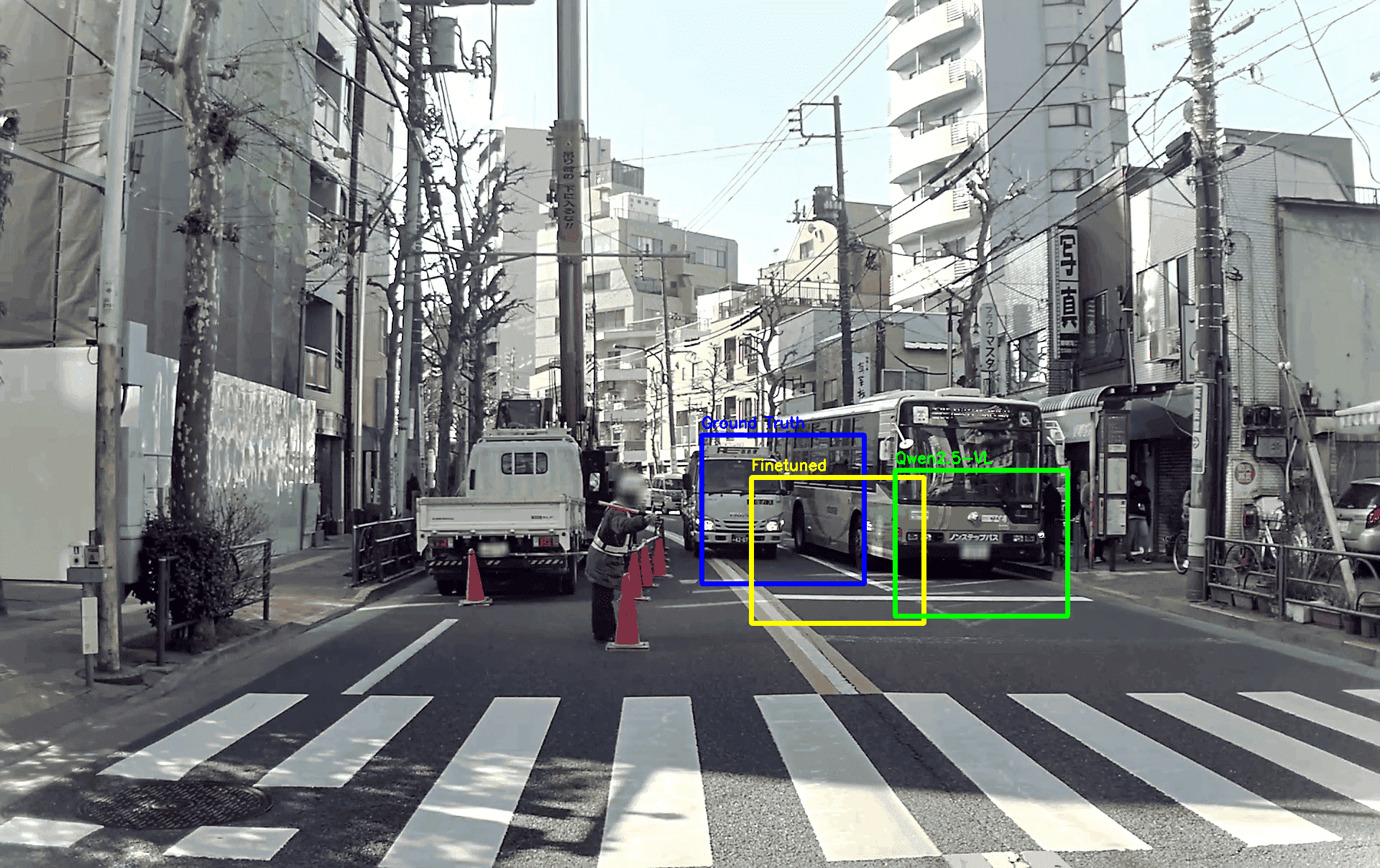}
            \end{tabular}
        \end{minipage} &
        % ---------- Fourth image ----------
        \begin{minipage}{0.45\linewidth}
            \centering
            {\scriptsize \textbf{Finetuned:} Yield} \\[0.2em]
            \begin{tabular}{@{}c@{}}
                \rotatebox{90}{\parbox{2cm}{\centering\scriptsize \textbf{GT:} Yield}} 
                \includegraphics[width=0.95\linewidth]{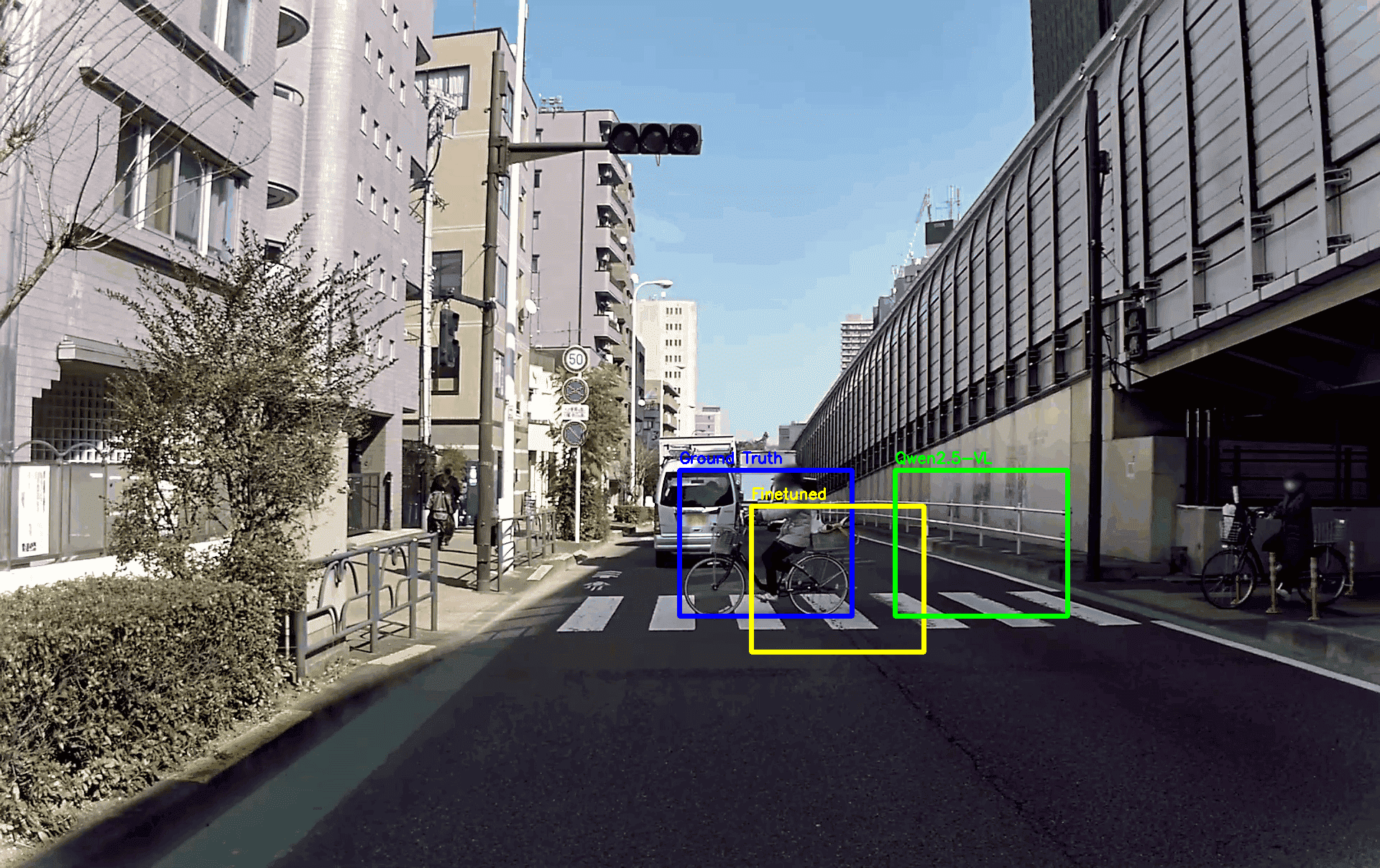}
            \end{tabular}
        \end{minipage}
    \end{tabular}
    \caption{ Qualitative comparison of \textbf{Safety Suggestions} across \colorbox{blue!30}{Ground-truth}, \colorbox{green!30}{Qwen2.5-VL} and \colorbox{yellow!30}{DriveSafe-Finetuned} in different challenging driving scenarios.}
    \label{fig:qual_safety_suggestions}
    \vspace{-0.3cm} 
\end{figure}

Fig.~\ref{fig:qual_safety_suggestions} highlights two common error types in Qwen2.5-VL and how DriveSafe-Finetuned overcomes them. The first is missing critical agents, where Qwen2.5-VL overlooks cyclists or pedestrians and generates irrelevant advice (e.g., ``\textit{follow the vehicle ahead}"), while DriveSafe provides accurate suggestions such as ``\textit{Slow down}" or ``\textit{Yield}", consistent with the ground-truth. The second is overcautious responses, where Qwen2.5-VL issues premature guidance (e.g., ``\textit{slow down}" when the leading vehicle is distant). DriveSafe instead produces balanced, context-aware advice (e.g., ``\textit{Be aware / cautious}"). These cases show that DriveSafe reduces ambiguity and delivers grounded, norm-consistent safety suggestions.

%Fig.~\ref{fig:qual_safety_suggestions} illustrates two major categories of errors commonly made by Qwen2.5-VL and shows how DriveSafe-Finetuned effectively addresses them. The first involves overlooking critical agents, where Qwen2.5-VL fails to recognize vulnerable road users such as cyclists or pedestrians and instead generates irrelevant advice like “follow the vehicle ahead,” whereas DriveSafe produces contextually accurate actions such as “Slow down” or “Yield,” aligning with the ground truth. The second failure mode arises from premature or exaggerated responses, where Qwen2.5-VL issues overly cautious guidance such as “slow down” even when the leading vehicle is far away, resulting in advice that is miscalibrated to the situation. In contrast, DriveSafe offers balanced and context-sensitive responses such as “Be aware / cautious,” avoiding both omission and overreaction. Together, these examples demonstrate DriveSafe’s ability to resolve ambiguity in safety-critical scenes and deliver grounded, precise, and norm-consistent safety suggestions.

\subsection{Ablation Experiments}

\noindent \textbf{Effects of Model Selection for Pseudo-labeling.}
We investigate the impact of different backbone models for generating pseudo-labels. For VLMs, pseudo-labels are derived using both the video and its ground-truth caption as input, while for LLMs, only the caption is provided. Table~\ref{tab:pseudo_ablation} reports performance on the safety suggestion prediction task. Among the compared models, LLaMA $3.1$~\cite{grattafiori2024llama3herdmodels} achieves the best $F_1$ score ($55.9$), outperforming all baselines. In contrast, DriveSafe-Finetuned attains a higher accuracy ($52.85$) but a substantially lower $F_1$ ($37.15$), suggesting a tendency toward majority-class predictions.

% Among the compared models, LLaMA 3.1~\cite{grattafiori2024llama3herdmodels} achieves the best overall results, yielding higher accuracy and $F_1$ scores compared to other models. Specifically, LLaMA 3.1 improves accuracy by 3.4\% and $F_1$ by 13.1\% over the strongest baseline, underscoring its effectiveness as a pseudo-labeling backbone.

\begin{table}[h]
\centering
\setlength{\tabcolsep}{6pt}
\renewcommand{\arraystretch}{1.15}
\begin{tabular}{lcc}
\toprule
\textbf{Pseudo-labeling Model} & \textbf{Accuracy} & \textbf{$F_1$} \\
\midrule
LLaVA-NeXT~\cite{li2024llavanextinterleavetacklingmultiimagevideo}      & 33.5 & 31.6 \\
Qwen2.5-VL~\cite{bai2025qwen25vltechnicalreport}        & 34.4 & 30.1 \\
DeepSeek~\cite{deepseekai2025deepseekv3technicalreport} & 44.2 & 42.8 \\
LLaMA-3.1-8B~\cite{grattafiori2024llama3herdmodels}  & 47.6 & \textbf{55.9} \\
\bottomrule
\textbf{DriveSafe-Finetuned} & \textbf{52.85} & 37.15 \\
\bottomrule
\end{tabular}
\caption{
Comparison of different models used for pseudo-label assignment, evaluated on safety suggestion prediction.}
\label{tab:pseudo_ablation}
\vspace{-.3cm}
\end{table}

\noindent \textbf{Effects of Contextual Cues.}
Table~\ref{tab:drama_ablation_caption} presents a component-level ablation study on the DRAMA~\cite{malla2023drama} dataset, where we progressively introduce three contextual cues. The baseline system, without any contextual cues, yields relatively low scores. Incorporating either spatial context $\mathcal{S}_t$ with depth $\mathcal{D}_t$, or motion $\mathcal{M}_t$ with depth $\mathcal{D}_t$, results in moderate gains METEOR~\cite{banerjee2005meteor} (+21.4\%) and (+27.7\%) respectively, highlighting the individual contributions of spatial and motion context. Combining both spatial $\mathcal{S}_t$ and motion $\mathcal{M}_t$ leads to larger gains (METEOR~\cite{banerjee2005meteor}: +41.03\%, CLAIR~\cite{chan2023clairevaluatingimagecaptions}: 8.88\%). Finally, the full model with all three contexts achieves the best results (METEOR~\cite{banerjee2005meteor}: 33.92, CLAIR~\cite{chan2023clairevaluatingimagecaptions}: 30.47), marking an 81.2\% improvement on METEOR and 43.7\% on CLAIR over the baseline.

\begin{table}[h]
\centering
\setlength{\tabcolsep}{6pt}
\renewcommand{\arraystretch}{1.15}
\begin{tabular}{ccccc}
\toprule
\textbf{$\mathcal{M}_t$} & \textbf{$\mathcal{S}_t$} & \textbf{$\mathcal{D}_t$} & \textbf{METEOR} & \textbf{CLAIR} \\
\midrule
\xmark & \xmark & \xmark & 18.72 & 21.19 \\
\xmark & \cmark & \cmark & 22.73 ($\uparrow$21.42\%) & 24.46 ($\uparrow$15.45\%) \\
\cmark & \xmark & \cmark & 23.90 ($\uparrow$27.67\%) & 21.78 ($\uparrow$2.81\%) \\
\cmark & \cmark & \xmark & 26.40 ($\uparrow$41.03\%) & 23.06 ($\uparrow$8.88\%) \\
\cmark & \cmark & \cmark & \textbf{33.92} ($\uparrow$81.20\%) & \textbf{30.47} ($\uparrow$43.75\%) \\
\bottomrule
\end{tabular}
\caption{\textbf{Component-level ablation on DRAMA~\cite{malla2023drama} dataset.} Relative improvements are computed w.r.t. the baseline (first row).}
\label{tab:drama_ablation_caption}
\vspace{-.2cm}
\end{table}

\begin{table}[h]
\centering
\setlength{\tabcolsep}{4pt}
\renewcommand{\arraystretch}{1.2}
\begin{tabular}{lccc}
\toprule
\textbf{Model} & \textbf{Params} & \textbf{METEOR$\uparrow$} & \textbf{CLAIR$\uparrow$} \\
\midrule
\multicolumn{4}{l}{\textit{VLM-wise comparison (LLaMA-3.1~\cite{grattafiori2024llama3herdmodels} fixed)}} \\
\midrule
LLaVA-Next Video~\cite{li2024llavanextinterleavetacklingmultiimagevideo} & 7B & 23.64 & 23.22\\
Qwen2.5 VL~\cite{bai2025qwen25vltechnicalreport} & 7B & \textbf{33.92} & 30.47 \\
\midrule
\multicolumn{4}{l}{\textit{LLM-wise comparison (Qwen2.5-VL~\cite{bai2025qwen25vltechnicalreport} fixed)}} \\
\midrule
% Gemini2.5-Pro~\cite{qwen2.5} & 7B &  & \\
DeepSeek~\cite{deepseekai2025deepseekv3technicalreport} & 7B & 14.66 & 34.46\\
LLaMA-3.1~\cite{grattafiori2024llama3herdmodels} & 8B & 33.92 & 30.47 \\
\bottomrule
\end{tabular}
\caption{DriveSafe model performance with different VLMs and LLMs.}
\label{tab:drivesafe_vlm_llm}
\vspace{-.2cm}
\end{table}

\noindent\textbf{Backbone Comparison.}
Table~\ref{tab:drivesafe_vlm_llm} presents an ablation study comparing different VLM–LLM backbones within DriveSafe. In the VLM-wise comparison (LLM fixed to LLaMA-3.1~\cite{grattafiori2024llama3herdmodels}), Qwen2.5-VL~\cite{bai2025qwen25vltechnicalreport} achieves approximately 40\% higher METEOR and 30\% higher CLAIR scores than LLaVA-NeXT~\cite{li2024llavanextinterleavetacklingmultiimagevideo}, underscoring its stronger temporal grounding and video understanding. In the LLM-wise comparison (VLM fixed to Qwen2.5-VL), LLaMA-3.1 provides consistently strong alignment, while DeepSeek~\cite{deepseekai2025deepseekv3technicalreport} yields a 55\% lower METEOR but a 13\% higher CLAIR score. Since CLAIR~\cite{chan2023clairevaluatingimagecaptions} relies on DeepSeek as its evaluator, this introduces a mild bias favoring its outputs. Overall, these results confirm that both VLM and LLM choices substantially influence DriveSafe’s performance, and that the optimal configuration requires balancing fine-grained caption accuracy with semantic risk-awareness.

\subsection{Application of DriveSafe}

\begin{figure}[t]
    \centering
    \includegraphics[width=1\linewidth]{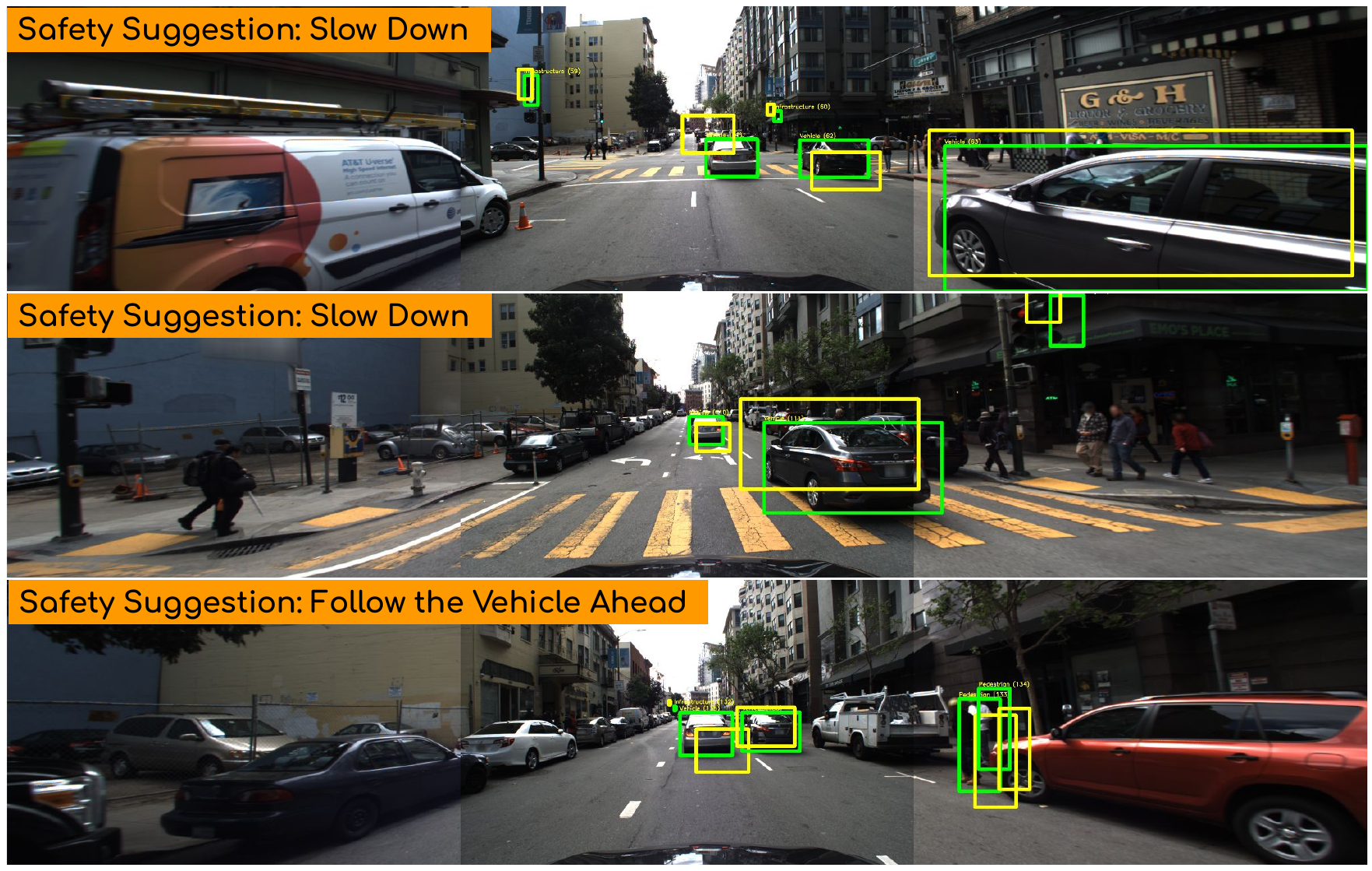}
    \caption{Top-to-bottom sequence with \colorbox{yellow!30}{DriveSafe-Finetuned} and \colorbox{green!30}{Ground-Truth} predictions; safety suggestions appear top-left.}
\label{fig:video_frames}
\vspace{-.5cm}
\end{figure}

We evaluate our model’s end-to-end ability to identify and localize multiple risky objects and generate safety suggestions in complex driving environments. For this evaluation, we use the Rank2Tell~\cite{sachdeva2023rank2tellmultimodaldrivingdataset} dataset, which is ideal due to its multi-object annotations across entire video sequences. DriveSafe generates multimodal contextual cues (spatial, depth, motion), converts them into captions, and uses these for downstream risk identification. Figure \ref{fig:video_frames} illustrates DriveSafe’s performance across sequential frames. The model consistently localizes a potentially risky vehicle with bounding boxes and provides context-aware safety suggestions that evolve with the scene: it recommends “\textit{Slow down}” as the car approaches the intersection (top and middle frames), and updates to “\textit{Follow the vehicle ahead}” once the situation stabilizes (bottom frame).

% We also generate outputs from Qwen2.5-VL for the same Rank2Tell video and compare them quantitatively against DriveSafe-Finetuned. Qwen2.5-VL achieves only 10.2% risk detection accuracy with an Acc@0.5 of 0.3, indicating limited ability to localize and prioritize risky objects. In contrast, DriveSafe-Finetuned demonstrates substantial gains, reaching 68.34% accuracy and 70.41 Acc@0.5, highlighting the effectiveness of grounding risk reasoning in multimodal captions.

\section{Conclusions}

This work introduced DriveSafe, a caption-based framework that enhances risk assessment in autonomous driving scenarios. Our evaluations demonstrate significant improvements over zero-shot MLLM baselines in risk assessment tasks, with fine-tuning substantially reducing hallucinations while improving the accurate identification and precise localization of risky driving behaviors. Beyond risk assessment, DriveSafe excels in safety applications by generating grounded safety recommendations that are directly linked to their underlying risk-inducing behaviors. This explicit connection between identified risks and corresponding safety suggestions addresses critical gaps in existing methods and provides the transparency necessary for building trust in autonomous systems. While the current risk-to-safety mapping is static, serving as a foundational first step to this novel framework. Future work will focus on learning dynamic mappings to improve scalability and adaptability, along with advanced environmental scaling, long-horizon temporal reasoning, and human-aligned explanation evaluation.

% Future work will target environmental scaling, long-horizon temporal reasoning, and human-aligned explanation evaluation.
%\newline
\textbf{Acknowledgment.} This project was supported by iHub-Data and Mobility at IIIT Hyderabad.

% \input{sec/X_suppl}
% \bibliographystyle{ACM-Reference-Format}
% \bibliography{main}
% \bibliographystyle{IEEEtranBST}
\bibliographystyle{IEEEtran}
\vspace{-.25cm}
\bibliography{reference}

\end{document}